\documentclass[11pt]{article}

\usepackage[]{EMNLP2023}

\usepackage{times}
\usepackage{latexsym}

\usepackage[T1]{fontenc}

\usepackage[utf8]{inputenc}

\usepackage{microtype}

\usepackage{inconsolata}
\usepackage{microtype}
\usepackage{microtype}
\usepackage{times}
\usepackage{latexsym}
\usepackage{amssymb}
\usepackage{amsfonts}
\usepackage{amsmath} 
\usepackage{amsthm} 
\usepackage{booktabs}
\usepackage{enumerate}
\usepackage{enumitem}
\usepackage{graphicx}
\usepackage{subfigure}
\usepackage{xspace}
\usepackage{float}
\usepackage{bbm}
\usepackage{bm}
\usepackage{multirow}
\usepackage{booktabs}
\usepackage{color}
\usepackage{framed}
\usepackage{stfloats}
\usepackage{iitem}
\usepackage{makecell}
\usepackage{float}
\usepackage{placeins}
\usepackage{afterpage}
\usepackage{bbding}
\usepackage[normalem]{ulem}
\usepackage{soul}
\usepackage{xcolor}
\usepackage{color}
\usepackage{colortbl}
\usepackage{algorithm}
\usepackage{algorithmic}
\usepackage{array}
\usepackage{utfsym}
\usepackage{hyperref}
\usepackage{listings}
\usepackage{lipsum,mwe,cuted}

\usepackage{url}
\newcommand{\paratitle}[1]{\vspace{1.5ex}\noindent\textbf{#1}}
\newcommand{\ie}{\emph{i.e.,}\xspace}

\newcommand{\eg}{\emph{e.g.,}\xspace}

\newcommand{\ignore}[1]{}

\usepackage{xcolor}
\definecolor{gold}{RGB}{205,133,63}
\definecolor{fGreen}{RGB}{34,139,34}
\definecolor{tOrange}{RGB}{255,165,0}
\definecolor{tBlue}{RGB}{135,206,250}
\definecolor{tPink}{RGB}{255,204,204}
\definecolor{tGreen}{RGB}{205,230,199}
\definecolor{tGold}{RGB}{255,215,0}
\newcolumntype{P}[1]{>{\centering\arraybackslash}p{#1}}
\newcolumntype{M}[1]{>{\centering\arraybackslash}m{#1}}

\usepackage{inconsolata}

\title{Small Agent Can Also Rock! Empowering Small Language Models as Hallucination Detector}

\author{
	Xiaoxue Cheng\textsuperscript{\rm{1}}, 
	Junyi Li\textsuperscript{\rm{1,3}},
        Wayne Xin Zhao\textsuperscript{\rm{1}\thanks{\ \ Corresponding author}\ },
        Hongzhi Zhang\textsuperscript{\rm{4}}, \\
        \textbf{Fuzheng Zhang\textsuperscript{\rm{4}}, 
        Di Zhang\textsuperscript{\rm{4}}, 
        Kun Gai\textsuperscript{\rm{4}} {\rm and} 
	Ji-Rong Wen\textsuperscript{\rm{1,2}}} \\
	\textsuperscript{1}Gaoling School of Artificial Intelligence, Renmin University of China \\
	\textsuperscript{2}School of Information, Renmin University of China \\
	\textsuperscript{3}DIRO, Universit\'{e} de Montr\'{e}al 
        \textsuperscript{4}Kuaishou \\
	\texttt{chengxiaoxue@ruc.edu.cn} \quad
        \texttt{lijunyi@ruc.edu.cn} \quad
	\texttt{batmanfly@gmail.com} \\
}

\begin{document}
\maketitle
\begin{abstract}
Hallucination detection is a challenging task for large language models (LLMs), and existing studies heavily rely on powerful closed-source LLMs such as GPT-4. In this paper, we propose an autonomous LLM-based agent framework, called \textbf{HaluAgent}, which enables relatively smaller LLMs\footnote{~``smaller'' in this work is relative to larger language models (\eg over 100B).} (\eg Baichuan2-Chat 7B) to actively select suitable tools for detecting multiple hallucination types such as text, code, and mathematical expression. In HaluAgent, we integrate the LLM, multi-functional toolbox, and design a fine-grained three-stage detection framework along with memory mechanism. 
To facilitate the effectiveness of HaluAgent, we leverage existing Chinese and English datasets to synthesize detection trajectories for fine-tuning, which endows HaluAgent with the capability for bilingual hallucination detection.
Extensive experiments demonstrate that only using 2K samples for tuning LLMs, HaluAgent can perform hallucination detection on various types of tasks and datasets, achieving performance comparable to or even higher than GPT-4 without tool enhancements on both in-domain and out-of-domain datasets.
We release our dataset and code at \url{https://github.com/RUCAIBox/HaluAgent}.
\end{abstract}

\section{Introduction}

Recently, large language models (LLMs)~\cite{zhao2023survey} have demonstrated exceptional capabilities across a variety of tasks within the field of natural language processing. However, \emph{hallucination}~\cite{ji2023survey, rawte2023survey, zhang2023siren, huang2023survey, ye2023cognitive} in text generated by LLMs remains an underlying concern, impeding the application of LLMs in real-world scenarios~\cite{kaddour2023challenges}.
\citet{xu2024hallucination} have indicated that, despite the existence of some effective hallucination mitigation strategies, the occurrence of hallucinations in LLMs is inevitable. Therefore, reliable and effective hallucination detection methods are necessary and urgent.

Existing hallucination detection methods can be roughly categorized into two primary approaches. 
One line of work relies on the internal knowledge of LLMs to directly identify hallucinations via prompts~\cite{li2023halueval, lei2023chain} or evaluating the semantic consistency among multiple responses to the same question generated by LLMs~\cite{manakul2023selfcheckgpt}. 
However, these methods are usually constrained not only by LLMs' internal knowledge but also by their abilities to utilize knowledge.
Another line of work extends the detection ability of LLMs by employing external tools (\eg search engine) to obtain supporting evidence for hallucination detection~\cite{chern2023factool, wei2024long, min2023factscore}. However, these approaches mostly depend on closed-source powerful LLMs such as GPT-4. Moreover, their detection process is usually pre-determined by human, making it difficult for the model to autonomously and effectively execute the hallucination detection. 

To address these issues, in this paper, we propose the \textbf{HaluAgent}, an autonomous LLM-based agent framework for hallucination detection. This agent is based on smaller open-source models (\ie Baichuan2-Chat 7B and 13B~\cite{yang2023baichuan}) and capable of \emph{bilingual} hallucination detection in Chinese and English. The motivations are twofold: (1) designing autonomous detection agents that can actively make decisions and judgements, without human
assistance; (2) enabling relatively smaller models to effectively perform complex detection, without reliance on close-sourced LLM APIs. To achieve this, we make three major technical contributions. First, we extend the LLM's capability to detect a broader range of hallucination forms such as text, code, math expression, or their combination by curating a multi-functional toolbox, in contrast to previous work only focused on textual hallucination or limited tools~\cite{manakul2023selfcheckgpt,min2023factscore}. Second, we design a fine-grained three-stage detection framework along with memory mechanism, including sentence segmentation, tool selection and verification, and reflection. Third, we leverage existing hallucination datasets to synthesize detection trajectories for fine-tuning the LLM, where we first employ GPT-4 to execute the above three stages until obtaining detection results consistent with the ground-truth label and then synthesize the instruction data. We compare HaluAgent and previous work in Table~\ref{tab: detection-methods}.

To verify the effectiveness, We evaluate HaluAgent on both in-domain and out-of-domain datasets at response- and sentence-level granularities. After fine-tuning with only 2K trajectories, the detection performance of HaluAgent has been improved significantly (\eg overall accuracy increases from 46.44\% to 79.70\% in four in-domain datasets and from 49.50\% to 78.43\% in two out-of-domain datasets), reaching a level comparable to or even higher than GPT-4 without tool enhancement. In the sentence-level detection experiments, HaluAgent also achieved substantial improvements, particularly with F1 scores on the math and science datasets increasing from 19.51\% to 68.80\% and from 17.54\% to 94.16\%, respectively.

\begin{table}[tb]
\centering
\small
\renewcommand\arraystretch{1.15}
\setlength\tabcolsep{2.0pt}
\resizebox{0.49\textwidth}{!}{
\begin{tabular}{l|ccccc}
\toprule
\textbf{Methods}  & \textbf{\makecell{Base\\Model}}& \textbf{\makecell{Task\\ Agnostic}} & \textbf{\makecell{Tool\\ Usage}} & \textbf{\makecell{Fine\\ Grained}} & \textbf{\makecell{Extensi-\\bility}}\\
\midrule
SelfCheckGPT & ChatGPT & \usym{2717} & \usym{2717} & \usym{2717} &  \usym{2717}\\
SAFE & GPT-4 & \usym{2717} & \usym{2713} & \usym{2713} & \usym{2717} \\
FacTool & GPT-4 & \usym{2717}  & \usym{2713} & \usym{2713} & \usym{2717}\\
HaluAgent & Baichuan2-Chat  & \usym{2713}  & \usym{2713} & \usym{2713} & \usym{2713} \\
\bottomrule
\end{tabular}
}
\caption{Comparison of different methods. \textbf{{Task Agnostic}} means whether the method is designed to specific tasks; \textbf{{Fine Grained}} describes whether providing detailed hallucination sentences; \textbf{{Extensibility}} means whether the method can extend to more tasks and tools.}
\label{tab: detection-methods}
\end{table}



\section{Approach}
\begin{figure*}[tb]
	\centering
	\includegraphics[width=1.00\textwidth]{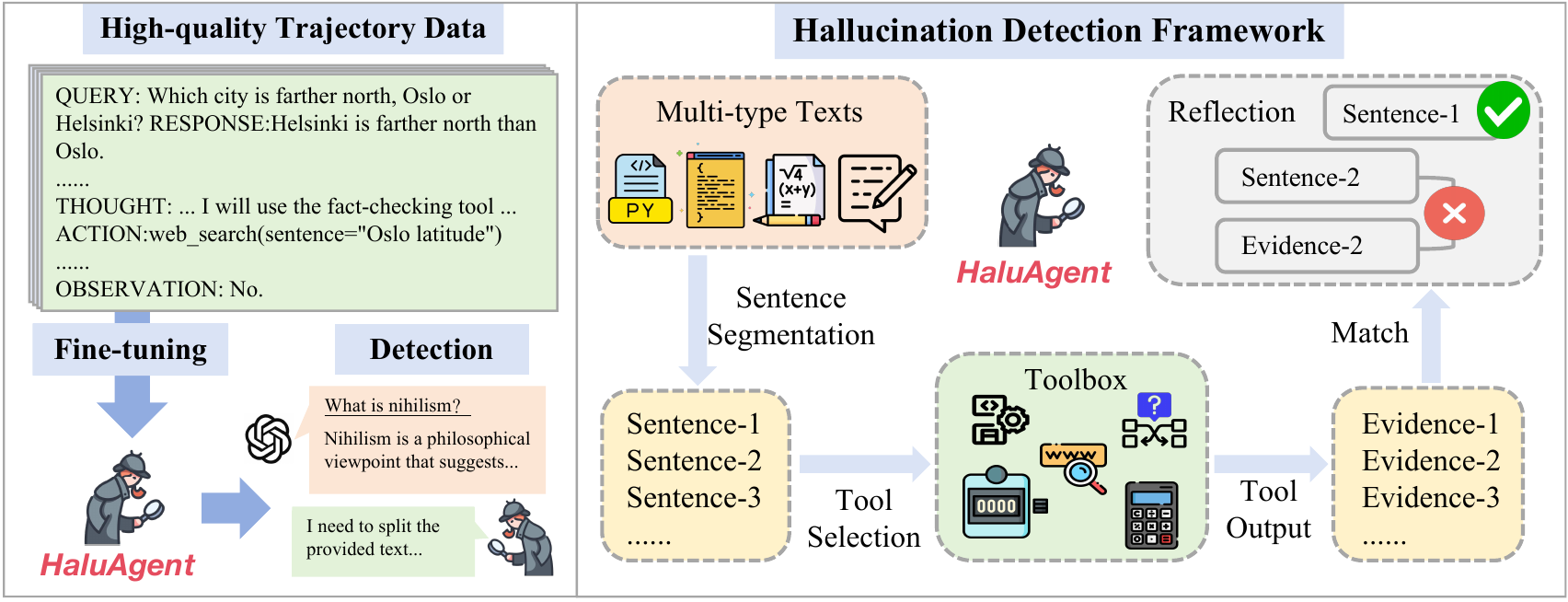}
	\caption{The overview of our proposed HaluAgent. The left part shows the process of fine-tuning open-source models and detecting hallucinations. The right part illustrates the hallucination detection pipeline of HaluAgent.}
	\label{fig: detection_framework}
	\vspace{-0.2cm}
\end{figure*}

In this section, we introduce \textbf{HaluAgent}, our proposed autonomous agent for detecting hallucinations across various text types. The core of our HaluAgent framework is a well-instructed LLM, which can autonomously leverage tools to detect a broader range of hallucination types. First, we define several tasks for hallucination detection and then design a toolbox with supporting tools to extend the LLM's capability. To enable step-by-step detection, we design a three-stage detection framework equipped with memory mechanism. Finally, we synthesize high-quality detection trajectory data to fine-tune open-source LLMs. We present the overall architecture of HaluAgent in Figure~\ref{fig: detection_framework}.

\subsection{Task Definition}
\label{sec: task}
Hallucination refers to seemingly plausible yet factually unsupported content~\cite{huang2023survey}. Unlike previous work that mostly focused on detecting text-based hallucinations~\cite{manakul2023selfcheckgpt,InterrogateLLM}, we consider a broader range of hallucination forms such as text, code, and mathematical expression.
Hence, in this work, we conduct hallucination detection in five tasks as:

\textbullet~\textbf{Knowledge-based QA} involves generating answers to the input question~\cite{kbqa}, and we aim to identify misinformation from the answer text such as incorrect historical dates, misattributed quotes, or false scientific facts.

\textbullet~\textbf{Conditional Text Generation} focuses on tasks with specific requirements given in the input instructions~\cite{pplm} such as generating text with particular length, format, or translating 
a paragraph into special language. We aim to detect hallucinations that deviate from the given instructions or include irrelevant information.

\textbullet~\textbf{Semantic Consistency} is not specific to any particular type of text or task. We define hallucination in semantic consistency as those irrelevant or self-contradictory content in the responses.

\textbullet~\textbf{Math Problem Solving} is related to generating a series of mathematical expressions to solve the math problem~\cite{hendrycks2021measuring}, while the expressions may contain computational errors, \eg arithmetic or calculation mistakes.

\textbullet~\textbf{Code Generation} aims to generate code snippets for the input query~\cite{chen2021evaluating}. We define hallucinations as code snippets that are syntactically incorrect, fail to execute as intended, or contain logical flaws and missing dependencies.

Note that in real-world scenarios the responses from LLMs may contain a mixture of these hallucination types, and we aim to develop a general and versatile hallucination detection agent that can deal with a broader range of hallucinations.

\subsection{Toolbox for Hallucination Detection}
Since detecting texts with a mix of hallucination types is challenging, we design a comprehensive toolbox to enhance the hallucination detection capability of LLMs following previous work~\cite{chern2023factool,gou2024critic}. Based on the hallucination types discussed in Section~\ref{sec: task}, we incorporate five types of external tools, \ie search engine, calculator, code interpreter, condition verifier, and semantic checker, and two internal system tools: 

\textbullet~\textbf{Search Engine} is utilized to retrieve supporting evidence from the web for identifying factually incorrect content, defined as \emph{web\_search}.
We use Google Programmable Search Engine API\footnote{https://developers.google.com/custom-search} to implement this tool.
Considering the relevance between retrieved documents and the output text, we only use the top-5 documents as the most relevant retrieval results for hallucination detection.

\textbullet~\textbf{Calculator} is used to verify inaccurate mathematical calculations in the model responses, targeting math-related hallucinations, defined as \emph{calculator}.
We implement calculator via the scientific computing library SymPy~\cite{10.7717/peerj-cs.103}.

\textbullet~\textbf{Code Interpreter} can be used to validate code snippets by executing the snippets in a programming environment, defined as \emph{code\_interpreter}, ensuring that the code is syntactically correct and functions as intended. 
We implement the code interpreter tool following CRITIC~\cite{gou2024critic}.

\textbullet~\textbf{Condition Verifier} aims to detect hallucinations for conditional text generation by assessing whether the text is consistent with the given condition. For example, we utilize word counter (\emph{word\_counter}) to compute the number of words in a length-constrained generation scenario.

\textbullet~\textbf{Semantic Checker} mainly addresses hallucination types such as irrelevant responses and self-contradictions, which is defined as \emph{match}. 
We leverage GPT-4 to examine the consistency and relevance of semantics, primarily handling the semantic matching scenario. 

\textbullet~\textbf{System Tools} are developed to support the basic manipulation operations for hallucination detection, including sentence segmentation (\emph{split\_text}) and returning detection results (\emph{get\_answer}).

All tools are defined as functions in a unified way and we present the whole toolbox in Table~\ref{tab: toolkit} in Appendix~\ref{app: toolkit}. It is worth noting that the toolbox is not limited to the existing tools and can be easily extended, \eg adding more task-specific tools.

\subsection{HaluAgent Framework}
\label{haluagent framework}

Inspired by prior work on complex reasoning~\cite{KGAgent}, we consider the hallucination detection process as an agent task. 
Specifically, HaluAgent includes three stages: sentence segmentation, tool selection, and reflection, along with memory mechanism. Below, we outline the workflow and components of our HaluAgent framework.

\paratitle{Sentence Segmentation.}
Since the responses of LLMs are usually the combination of facts, opinions, and various types of texts, we first segment the responses into several independent detection units. To be specific, we utilize the system tool, \textit{split\_text}, to perform sentence segmentation. This tool requires the agent to split the input text into a set of sentences and complete any incomplete semantic information within the sentences, \eg pronouns and omitted content.
Sentence segmentation reduces the complexity of hallucination detection tasks and allows HaluAgent to tailor its detection strategies for individual sentences, ensuring more accurate and fine-grained results by minimizing interference from unrelated content.

\paratitle{Tool Selection and Verification.}
Next, HaluAgent verifies each sentence separately by selecting the appropriate tool from the toolbox based on the type of sentence content (\eg \textit{web\_search} for fact statements and \textit{calculator} for mathematical expressions). HaluAgent compares each sentence with the execution results of the selected tool to identify any inconsistencies or inaccuracies, thereby detecting hallucinations.
To prevent the agent from forgetting intermediate detection results and support the subsequent reflection, HaluAgent stores the detection result of each sentence as a triple, \ie \emph{(sentence, hallucination label, supporting evidence)}. If the sentence is identified as containing hallucinations, it will be labeled ``1'', otherwise ``0''. These useful information will be stored based on a memory mechanism,
allowing HaluAgent to refer to historical results and maintain a consistent understanding of the text veracity throughout the detection process.

\paratitle{Reflection.}
After individually examining all the sentences, HaluAgent can obtain preliminary detection results. However, due to the limited capacity of each tool and potential errors in their outputs, these detection results might not be fully accurate. 
To address this, HaluAgent performs the final reflection to double-check whether the previous detection results are correct from \emph{local} and \emph{global} perspectives. At the local level, HaluAgent will match each sentence with the corresponding evidence to ensure the local correctness of hallucination detection. However, different sentences may influence each other. Therefore, at the global level, HaluAgent will determine whether the current sentence is incorrect based on the context of other sentences.
For instance, if the calculation result in a preceding sentence is incorrect, any subsequent steps based on this result should be considered incorrect, even if these steps are correct when checked in isolation.
Any detection mistakes will be corrected and the detection results (\ie hallucination label and supporting evidence) stored in memory are updated accordingly.
After reflection, HaluAgent invokes a system tool, \textit{get\_answer}, to output the final detection result. If any hallucinations are detected, HaluAgent outputs the specific sentences and supporting evidence from the detection tools.

Throughout the above process, HaluAgent first segments the input text into a set of semantically complete sentences, then selects tools to check each sentence individually, and finally reflects on the detection results to further correct mistakes. To support this process, we use memory mechanism to store useful information such as historical detection trajectories and current detection results.

\subsection{Bilingual Agent Tuning}

Previous studies mostly depended on closed-source LLMs (\eg GPT-4) to detect hallucinations~\cite{chern2023factool,wei2024long}. To empower smaller language models as effective hallucination detectors, we aim to perform supervised fine-tuning on smaller LLMs (\eg Baichuan2-Chat 7B). Given the powerful agent capability, we leverage GPT-4 to synthesize high-quality hallucination detection trajectory data in Chinese and English following the detection framework in Section~\ref{haluagent framework}.

\subsubsection{Trajectory Generation}

\begin{table}[tb]
\centering
\small
\begin{tabular}{l|ccc}
\toprule
Datasets  & \#Train & \#Trajectory & \#Test \\
\midrule
WebQA & 900 & 675 & 100 \\
Ape210K & 500 & 334 & 100 \\
HumanEval & 100 & 100 & 63 \\
WordCnt & 100 & 100 & 100 \\
HaluEval-QA & 900 & 808 & 100 \\
\midrule
All & 2500 & 2017 & 463 \\
\bottomrule
\end{tabular}
\caption{Statistics of synthetic detection trajectories.}
\vspace{-0.3cm}
\label{tab: trajectory statistics}
\end{table}

\paratitle{Data Source.}
\label{sec: in-domain dataset}
To curate high-quality trajectory data covering diverse hallucination types, we select and construct five datasets, \ie HaluEval~\cite{li2023halueval} and WebQA~\cite{li2016dataset} for knowledge-based QA, Ape210K~\cite{zhao2020ape210k} for math word problem, HumanEval~\cite{chen2021evaluating} for code generation, and WordCnt for conditional text generation. Among them, HaluEval and HumanEval are English datasets, and WebQA, Ape210K, and WordCnt are Chinese datasets, which enable bilingual detection capabilities of HaluAgent. In our experiments, we use GPT-4 to synthesize WordCnt which is targeted at generating text with a specified length. Besides, we obtain the ground-truth hallucination labels for WebQA and Ape210K by ChatGPT and human annotators.
We provide details of each dataset in Appendix~\ref{app: data_source}.

\paratitle{Trajectory Format.}
Based on our datasets, we employ GPT-4 to execute the detection process in Section~\ref{haluagent framework} and generate corresponding trajectory data following the ReAct format~\cite{yao2023react}. We begin by feeding the detection instruction and the text to be detected as input for GPT-4. At each turn, the agent receives an \emph{observation}, makes its plans and thoughts as \textit{thought}, and invokes corresponding tools through \textit{action}. The results from these tools are formulated as a new observation for the next turn. By iterating the above process, we can obtain a complete detection trajectory comprising the input instruction, detected text, intermediate steps (\ie observations, thoughts, actions), and the final detection result. To ensure the accuracy of the trajectories, we remove those samples that include wrong tool invocation, formatting errors, and inconsistency between the detection result and the ground-truth hallucination label. Finally, we produce 2,017 high-quality trajectories for supervised fine-tuning. We present an example in Figure~\ref{fig: case study} and the statistics of trajectory data in Table~\ref{tab: trajectory statistics}.

\subsubsection{Trajectory Tuning}
Based on the above formatted bilingual trajectory data, we perform supervised fine-tuning on Baichuan2-Chat 7B and 13B~\cite{yang2023baichuan}, which are much smaller than the backbone models in previous studies~\cite{chern2023factool,wei2024long}. 
Formally, the hallucination detection trajectory for each sample can be represented as $\langle o_0, t_1, a_1, o_1, t_2, a_2, \ldots, o_{n-1}, t_n, a_n, o_n \rangle$, where $o_i$, $t_i$, and $a_i$ denote the observation, {thought}, and {action} at the $i$-th turn, respectively. Specifically, $o_0$ denotes the initial observation consisting of the input instruction and detected text, and $o_n$ denotes the final detection result. At each turn, based on the historical trajectory $c_i = \langle o_0, t_1, a_1, \ldots, o_{i-1} \rangle$, the agent aims to generate thought $t_i$ and action $a_i$.
Therefore, during the trajectory fine-tuning process, we only compute the cross-entropy loss for $t_i$ and $a_i$ while masking $o_i$:
\begin{equation}
    \mathcal{L} = -\log \sum_{i=1}^{n} \text{Pr}(t_i,a_i|c_i).
\end{equation}

\subsection{Comparison to Previous Work}

To clarify the differences between HaluAgent and other hallucination detection methods, we aim to address the following two questions:

\textbullet~\textbf{What are the benefits of designing an autonomous hallucination detection agent?} Hallucination detection is fundamental to related research. Most previous work either relied on the internal knowledge of LLMs (might be limited)~\cite{LMvsLM} or performed coarse-grained detection process~\cite{manakul2023selfcheckgpt}. Designing an agent for hallucination detection offers a flexible alternative that can effectively extend the detection capability of LLMs through tool utilization.
Existing agent-based detection methods do not consider tool utilization or only employ limited tools such as search engine for hallucination detection in long texts~\cite{lei2023chain,wei2024long}.
In contrast, HaluAgent develops a comprehensive and extensible toolbox and performs fine-grained reasoning process for hallucination detection, providing a more adaptable and robust solution.

\textbullet~\textbf{Can smaller language models perform well in challenging hallucination detection?}
Existing methods~\cite{chern2023factool, manakul2023selfcheckgpt, dhuliawala2023chain} heavily depended on powerful closed-source LLMs such as GPT-4, which leads to high computational costs and poses unavoidable limitations for the practical deployment of these technologies. Moreover, relying on closed-source models makes the detection results difficult to reproduce. 
Our work demonstrates that by incorporating the agent capabilities and tool integration, smaller language models can also effectively handle challenging hallucination detection tasks.
This way can provide a more viable and economical choice, significantly reducing the need for closed-source LLMs.

\section{Experiments}

\subsection{Experimental Setup}

\paratitle{In-domain/Out-of-domain Tasks.}
We evaluate HaluAgent on both in-domain and out-of-domain datasets. As described in Section~\ref{sec: in-domain dataset}, we select HaluEval-QA, WebQA, Ape210K, HumanEval, and WordCnt as in-domain datasets. For out-of-domain datasets, we use a Chinese dataset, HalluQA~\cite{cheng2023evaluating}, and an English dataset, HaluEval 2.0~\cite{li2024dawn}, which cover diverse hallucination detection scenarios including knowledge, math, science texts. 
All datasets are associated with ground-truth response-level hallucination labels. We present the details in Appendix~\ref{app: ood_dataset}.

\paratitle{Sentence-level Tasks.}
HaluAgent performs fine-grained hallucination detection by segmenting sentences and provides sentence-level detection results. We use FacTool~\cite{chern2023factool} with annotated claim-level hallucination labels to evaluate the fine-grained detection capability. We consider each claim as a sentence and concatenate all claims as the input text. FacTool contains five sub-datasets: Chinese-QA, KB-QA, math problems, code generation, and scientific literature review. Since the code generation sub-dataset is collected from HumanEval and overlaps with our training data, we conduct sentence-level detection experiments on the other four sub-datasets. 

\paratitle{Baselines.}
Unlike previous work relied on powerful closed-source models like GPT-4, HaluAgent is built upon relatively smaller language model by performing fine-tuning on hallucination detection trajectory data.
Hence, we compare HaluAgent with two kinds of baselines: (1) Closed-source models: \emph{GPT-4 prompt} and \emph{GPT-4 pipeline} employ GPT-4 with simple task description prompts and our proposed detection pipeline, respectively; (2) Open-source models: \emph{Baichuan2-Chat (7B and 13B)}, which is the backbone model of HaluAgent. For other detection methods without comparison in this work, they are either implemented based on ChatGPT/GPT-4 (an unfair comparison) or focus solely on specific tasks, making it difficult to adapt to other tasks, as shown in Table~\ref{tab: detection-methods}.

\paratitle{Metrics.}
Hallucination detection is essentially a binary classification task. Consequently, we adopt \emph{Accuracy} and \emph{F1 score} as metrics for response-level detection evaluation. Considering the imbalanced hallucination data distribution at the sentence level, we adopt \emph{Accuracy}, \emph{Precision}, \emph{Recall}, and \emph{F1 score} for sentence-level detection.

\begin{table*}[htb]
\centering\small
\renewcommand{\arraystretch}{1.15}
\resizebox{\textwidth}{!}{
\begin{tabular}{@{}c@{\hspace{0.8\tabcolsep}}lcc|cc|cc@{}}
\toprule
\multicolumn{1}{c}{\multirow{2}{*}{Types}} & \multicolumn{1}{l}{\multirow{2}{*}{Datasets}} & \multicolumn{2}{c}{GPT-4} & \multicolumn{2}{c}{Baichuan2-Chat} & \multicolumn{2}{c}{HaluAgent} \\ \cmidrule(l){3-4} \cmidrule(l){5-6} \cmidrule(l){7-8} 
\multicolumn{1}{l}{} & \multicolumn{1}{c}{} & prompt & \multicolumn{1}{c}{pipeline} & 7B & 13B & 7B & 13B  \\ \midrule
\multirow{6}{*}{\begin{tabular}[c]{@{}c@{}}{In-domain}\\ {Datasets}\end{tabular}} & {WebQA} & 82.00/35.71 & \underline{91.00}/\underline{57.14} & 51.00/14.04 & 54.00/\textbf{61.67} & 80.00/54.55 & \textbf{82.83}/51.43  \\
 & {Ape210K} &72.33/74.21 & \underline{76.63}/\underline{75.10} & 49.00/7.27 & 51.33/58.29& 72.00/72.55& \textbf{73.40}/\textbf{73.68}  \\
 & {HumanEval} & 71.43/79.07& \underline{93.44}/\underline{94.12} & 34.92/49.38& 47.62/19.51 & \textbf{93.44}/\textbf{94.12} & \textbf{93.44}/\textbf{94.12} \\
 & {WordCnt} & 56.00/66.15 & \underline{100.00}/\underline{100.00} & 43.00/16.00 & 46.00/59.70 & \textbf{100.00}/\textbf{100.00} & \textbf{100.00}/\textbf{100.00} \\
 & {HaluEval-QA} & 62.00/42.42 & \underline{77.53}/\underline{75.61} & 53.19/46.34 & 60.00/67.74 & 67.00/67.33 & \textbf{71.00}/\textbf{72.38} \\\cmidrule{2-8}
 & {Overall} & 69.76/71.66 & \underline{85.10}/\underline{83.12} & 46.44/32.20 & 52.70/55.58 & 79.70/79.50 & \textbf{81.86}/\textbf{80.69} \\ \midrule
\multirow{3}{*}{\begin{tabular}[c]{@{}c@{}}{Out-of-domain}\\ {Datasets}\end{tabular}} & {HalluQA} & 61.00/74.84 & \underline{85.11}/\underline{89.23} &33.00/12.99 & 56.00/67.16 & 67.48/76.09 & \textbf{78.16}/\textbf{83.75}  \\
 & {HaluEval 2.0} & 63.00/74.13 & \underline{85.71}/\underline{87.36} & 54.00/69.33 & 43.00/46.73 & 75.00/76.19 & \textbf{79.00}/\textbf{78.79}  \\ \cmidrule{2-8}
 & {Overall} & 62.00/74.50& \underline{85.25}/\underline{88.76} & 43.50/50.22 & 49.50/58.10 & 69.97/76.12 & \textbf{78.43}/\textbf{82.45}   \\ 
 \bottomrule
\end{tabular}
}
\caption{Evaluation results at Accuracy and F1 score on in-domain and out-of-domain datasets. \textbf{Bold} denotes the best methods among open-source models; \underline{underline} denotes the best methods among closed-source models.}
\label{tab: main-results}
\end{table*}

\begin{table*}[htb]
\centering\small
\renewcommand{\arraystretch}{1.15}
\resizebox{\textwidth}{!}{
\begin{tabular}{lcccc}
\toprule
 Models & Chinese-QA & KB-QA & Math & Science \\
 \midrule
 GPT4-prompt & 54.93/59.40/33.17/42.57 & 75.97/39.29/50.00/44.00 & 55.27/49.25/23.74/32.04 & 59.14/64.71/81.82/72.26\\
 GPT4-pipeline &79.76/64.71/50.00/56.41 & 84.12/85.31/93.21/89.09 & 91.61/79.66/78.33/78.99 & 95.72/96.77/98.04/97.40\\
 Baichuan2-Chat 7B  & 65.15/15.02/29.17/19.83 & 49.77/38.78/19.59/26.03  & 68.07/17.91/21.43/19.51 & 24.19/9.80/83.33/17.54\\
 Baichuan2-Chat 13B & 56.82/21.70/22.67/22.17 & 65.41/30.77/30.00/30.38 & 70.97/13.85/20.93/16.67  & 18.82/1.31/100.00/2.58\\
 HaluAgent-7B  & 73.57/32.96/55.97/41.49  & 83.33/41.82/82.14/55.42 & 85.39/81.13/59.72/68.80 & 90.27/94.77/93.55/94.16\\
 HaluAgent-13B & 75.47/33.80/56.25/42.23 & 81.97/62.07/85.71/72.00 & 87.50/64.52/72.73/68.38 & 92.39/96.05/94.81/95.42\\
\bottomrule
\end{tabular}
}
\caption{Evaluation results of sentence-level detection on the four subsets of FacTool.}
\label{tab: sentence-level results}
\vspace{-0.2cm}
\end{table*}

\subsection{Response-level Detection}

We present the evaluation results on in-domain and out-of-domain datasets in Table~\ref{tab: main-results}.

First, by comparing GPT-4 and Baichuan2-Chat, we can observe a large performance gap between closed-source and open-source models when detecting hallucinations solely based on their internal knowledge. In Table~\ref{tab: main-results}, 
GPT-4 achieves detection accuracies of 82.00\% and 72.33\% on WebQA and Ape210K respectively, whereas Baichuan2-Chat only achieves 51.00\% and 49.00\%. 
This underscores the considerable difference in hallucination detection capabilities between these models.

Second, implementing a fine-grained detection framework and incorporating extensive tools can enhance the hallucination detection performance of LLMs across various tasks. Compared to GPT-4 prompt, GPT-4 pipeline consistently yields better detection results across all datasets. 
Especially for tasks that require tools to detect the hallucinations such as WordCnt, our tool-assistant framework boosts the detection accuracy of GPT-4 from 56.00\% to 100.00\%. This demonstrates that guiding the model to utilize appropriate tools tailored to the specific text is an effective strategy to enhance hallucination detection capabilities.

Finally, based on trajectory fine-tuning, smaller open-source models can be effective autonomous agents for hallucination detection, narrowing the gap with closed-source models. As can be observed from Table~\ref{tab: main-results}, HaluAgent consistently improves detection performance across in-domain datasets compared to Baichuan2-Chat. 
For instance, on the WebQA dataset, HaluAgent 7B and 13B improve the detection accuracy from 51.00\%, 60.00\% to 80.00\%, 71.00\%, respectively.
Furthermore, substantial performance improvements are observed on out-of-domain datasets, with accuracy increasing from 33.00\% to 67.48\% (7B) for HalluQA and from 43.00\% to 79.00\% (13B) for HaluEval 2.0.  
With tool utilization and agent capabilities, HaluAgent achieves performance comparable to or even higher than GPT-4 prompt across all tasks, showing strong generalization capability.

\subsection{Sentence-level Detection}

In our framework, HaluAgent is capable of detecting hallucinations for each individual sentence. Thus, we evaluate the accuracy of HaluAgent in identifying hallucinations at the sentence level. 

The sentence-level detection results are shown in Table~\ref{tab: sentence-level results}. As we can see, HaluAgent achieves much higher F1 score compared to Baichuan2-Chat, especially for those tasks where the detection results can be precisely judged via tools. For example, HaluAgent 13B achieves an accuracy of 87.50\% on the math dataset, and F1 score of 95.42\% on the scientific literature review dataset, approaching the performance of the GPT-4 pipeline. We attribute this improvement to the fine-grained design of the HaluAgent framework and its capability to store and reflect on detection results for each sentence.

\begin{figure}[tb]
	\centering
	\includegraphics[width=0.48\textwidth]{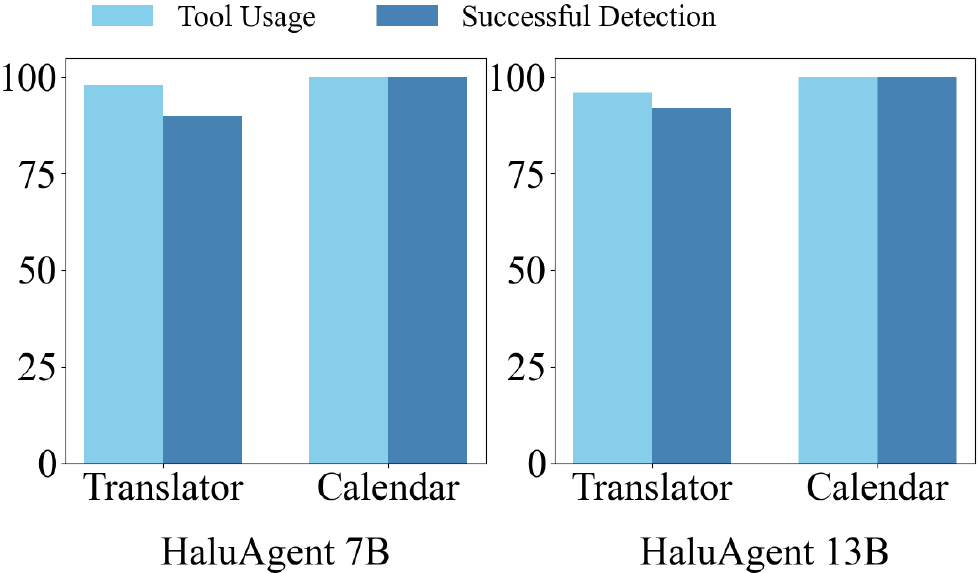}
	\caption{The usage rate of new tools and the proportion of successful detection.}
	\label{fig: new_tool}
	\vspace{-0.2cm}
\end{figure}

\begin{figure*}[tb]
	\centering
        \small
	\includegraphics[width=1.00\textwidth]{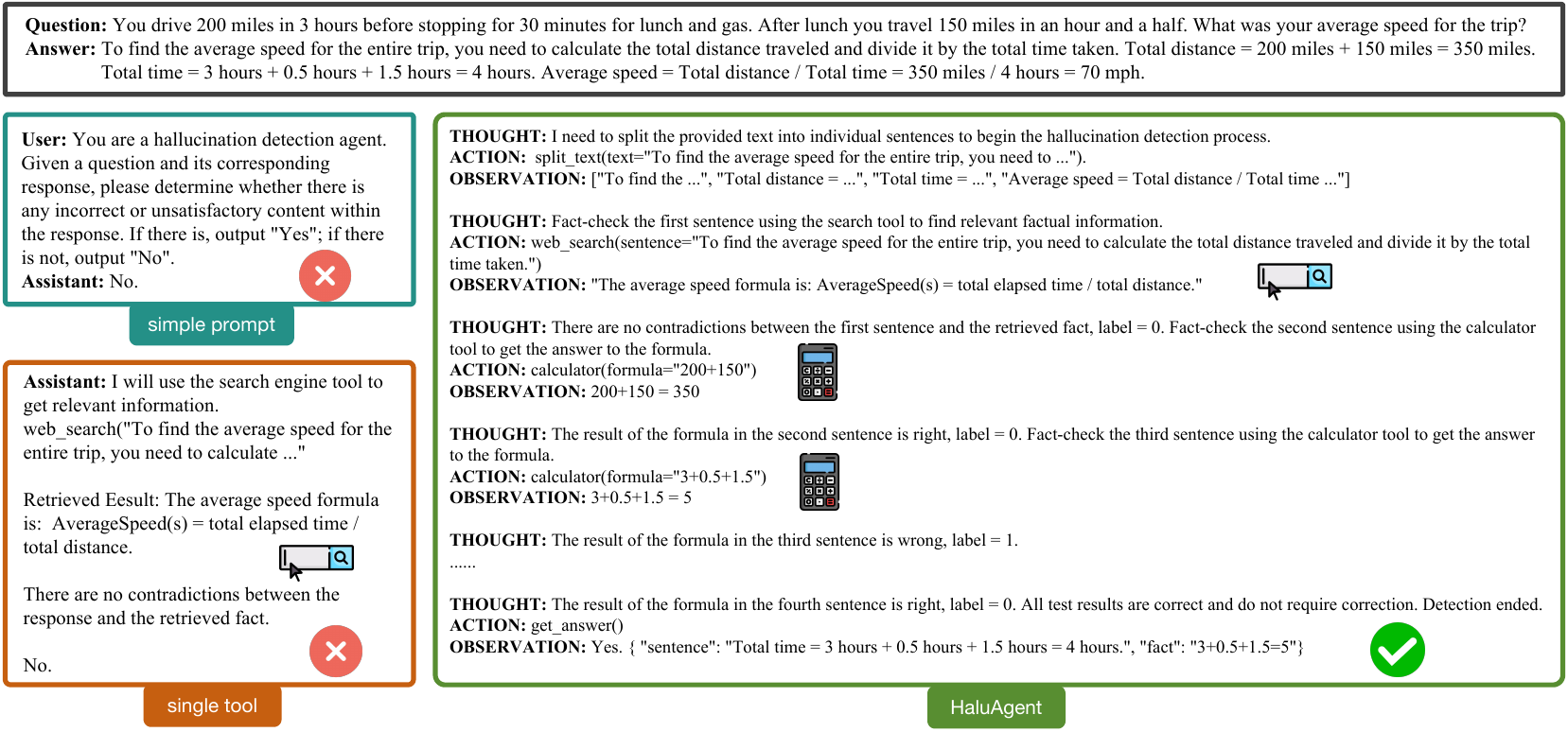}
	\caption{Case study between GPT-4 with a simple prompt and single tool, and the HaluAgent framework.}
	\label{fig: case study}
	\vspace{-0.2cm}
\end{figure*}

\subsection{Further Analysis}


\paratitle{Extensibility Study.}
To verify the extensibility of our HaluAgent framework, we introduce new tools to the fine-tuned models and test their abilities to use these tools for hallucination detection. Specifically, we incorporate a translator and a calendar tool into the HaluAgent toolbox for handling texts related to translation and date calculations. We provide instructions and two examples as in-context demonstrations to guide the model to use these new tools.
For evaluation, we employ ChatGPT to generate 100 samples involving translation and date calculation. Instruction and dataset details can be found in Appendix~\ref{app: scalability_exp}.

We measure the proportion of correct tool usage and successful task completion rate by HaluAgent in Figure~\ref{fig: new_tool}. As we can see, HaluAgent 7B and 13B achieve a usage rate of over 95\% for the translator tool, and the usage rate for the calendar tool is 100\%. Due to the models' inherent multilingual capabilities, they sometimes directly assess the accuracy of the translation result instead of invoking the translator tool. With the assistance of the appropriate tools, HaluAgent achieves a success rate of 100\% in hallucination detection for date calculation and over 90\% success rate for translation task. These experimental results indicate that with instructions and demonstrations, HaluAgent can effectively use new tools to complete hallucination detection tasks without additional fine-tuning. 

\paratitle{Case Study.}
To qualitatively demonstrate the effectiveness of HaluAgent, we present a case study about average speed calculation, which involves both commonsense knowledge and mathematical calculations. 
We compare HaluAgent with two hallucination detection methods based on GPT-4 with simple prompts and search engine tool, as shown in Figure~\ref{fig: case study}. 
As we can see, GPT-4 fails to detect the minor calculation errors (\ie \textit{3 hours + 0.5 hours + 1.5 hours = 4 hours}) when only provided with a simple prompt of task description. 
Meanwhile, GPT-4 with a search engine tool can only verify the formula for average speed in the response but cannot check the correctness of each calculation step. Consequently, both methods yield incorrect detection results.
In contrast, HaluAgent autonomously plans and selects suitable tools for different parts of the text (\ie search engine for knowledge checks and calculator for calculation checks), allowing for accurate verification of the content. This highlights HaluAgent's capability to effectively verify complex texts with mixed types of hallucinations.

\section{Related Work}

\paratitle{Hallucination Detection.}
Hallucination Detection in LLMs is a pivotal concern due to its role in identifying inaccuracies or falsehoods within model responses. Most existing methods for hallucination detection are implemented based on powerful LLMs~\cite{luo2024hallucination}. One category of these methods relies on the internal knowledge and consistency of LLMs to detect hallucinations, by breaking down the detection process~\cite{dhuliawala2023chain} or comparing multiple responses to the same query~\cite{manakul2023selfcheckgpt}. 
However, these approaches are inherently limited by the knowledge boundaries of LLMs. Another category of methods leverages external tools for hallucination detection~\cite{chern2023factool, wei2024long}. 
While tool utilization can compensate for the knowledge limitations of LLMs, existing methods often require manually designing specific tools tailored to particular text types. In contrast, HaluAgent is equipped with a versatile and expandable toolbox, enabling small language models to autonomously select appropriate tools for fine-grained detection.

\paratitle{Agent Tuning.}
Recently, the surprising planning and reasoning capabilities of LLMs have inspired research into their application as agents for specific tasks. 
Previous research~\cite{yao2023react, nakano2021webgpt, singh2023progprompt, yang2023mm} has predominantly relied on prompting to use LLMs as agents, but this method demands advanced instruction-following capabilities that open-source LLMs typically do not match with API-based LLMs.
To break this restriction, AgentTuning~\cite{zeng2023agenttuning} first fine-tunes open-source LLMs on agent interaction trajectories generated by powerful LLMs. Moreover, Agent-FLAN~\cite{chen2024agent} goes a step further by decomposing and redesigning the training corpus.
Besides general agents, KG-Agent~\cite{KGAgent} fine-tunes LLaMA-7B model to achieve an agent specialized in reasoning over knowledge graphs. Similarly, HaluAgent is fine-tuned on trajectory data of hallucination detection tasks, thereby enhancing the detection capabilities of open-source LLMs.
\section{Conclusion}
In this work, we proposed an autonomous agent framework, HaluAgent, which is capable of bilingual hallucination detection in Chinese and English. In our approach, we first curated a multi-functional toolbox to extend the LLM’s capability to detect hallucinations. Next, we designed a fine-grained three-stage detection framework along with memory mechanism, including sentence segmentation, tool selection and verification, and reflection.
Then, we leveraged existing datasets to synthesize detection trajectories by employing GPT-4 to execute the detection process following the HaluAgent framework.
Finally, we fine-tuned Baichuan2-Chat 7B and 13B on the synthesized trajectories. The HaluAgent models achieved notable improvements on both in-domain and out-of-domain datasets, with performance comparable to or even higher than GPT-4 without tool enhancements. 
Due to its high flexibility and adaptability, HaluAgent can effectively serve as a hallucination detector when human users interact with LLMs in real-world scenarios. In future work, we will extend our method to deal with more types of tools and hallucinations.

\section*{Limitations}
Although HaluAgent significantly enhances the performance of small open-source models in hallucination detection tasks, our approach still has some limitations. 
First, we use only Baichuan2-Chat as the backbone LLM and do not compare it with other models of comparable parameter size, such as Llama2-7B~\cite{touvron2023llama}, Mistral-7B~\cite{jiang2023mistral}, and Qwen-7B~\cite{qwen}. 
Second, our work focuses on hallucination detection tasks and does not propose corresponding mitigation strategies.
Third, we focus on detecting hallucinations that contain errors and contradictions, lacking consideration for hallucinations related to identity recognition and ethical issues.
Finally, the training process of HaluAgent uses only correct trajectory data for supervised fine-tuning, without fully leveraging failed trajectory data or incorporating other types of data. 
In the future, we will further refine the HaluAgent framework to cover more hallucination types and fully leverage failed data to train the model. We also consider developing improved mitigation strategies based on HaluAgent.

\bibliography{emnlp2023}
\bibliographystyle{acl_natbib}

\newpage
\appendix

\section{HaluAgent Framework}
\label{app: instructions}
\subsection{Instructions}

During the trajectory generation phase, we provide detailed instructions for the HaluAgent framework to prompt GPT-4 to generate the required detection trajectories. The English and Chinese prompts are shown in Figure~\ref{fig: english_instruction} and Figure~\ref{fig: chinese_instruction}, respectively.

\subsection{Toolbox}
\label{app: toolkit}

HaluAgent includes both verification tools and system tools. The tool names and their usage instructions are summarized in Table~\ref{tab: toolkit}.

\begin{table*}[b]
\centering
\renewcommand{\arraystretch}{1.10}
\resizebox{\textwidth}{!}{
\begin{tabular}{@{}c l p{10cm} @{}}
\toprule
\textbf{Category} & \textbf{Tool Name} & \textbf{Tool Usage Instruction} \\
\midrule
\multirow{14}{*}{\begin{tabular}[c]{@{}c@{}}\textbf{Verification}\\ \textbf{Tools}\end{tabular}} 
& web\_search & \textit{Input:} sentence: str $\rightarrow$ \textit{Output:} fact \\
& & Conduct a web search and return factual information. \\
\cmidrule(l){2-3}
& calculator & \textit{Input:} formula: str $\rightarrow$ \textit{Output:} result \\
& & Perform calculations based on the input formula and return the result. \\
\cmidrule(l){2-3}
& code\_interpreter & \textit{Input:} code: str $\rightarrow$ \textit{Output:} label \\
& & Execute code and return a label indicating whether the execution was successful. \\
\cmidrule(l){2-3}
& word\_count & \textit{Input:} length: int, text: str $\rightarrow$ \textit{Output:} count, label \\
& & Count words in the text and provide a label indicating whether the requirement is met. \\
\cmidrule(l){2-3}
& match & \textit{Input:} sentence: str, context: str $\rightarrow$ \textit{Output:} label \\
& & Match a sentence with the given context and return a label indicating whether semantic matching is successful. \\
\midrule
\multirow{4}{*}{\begin{tabular}[c]{@{}c@{}}\textbf{System}\\ \textbf{Tools}\end{tabular}} 
& split\_text & \textit{Input:} text: str $\rightarrow$ \textit{Output:} sentences \\
& & Split text into individual sentences. \\
\cmidrule(l){2-3}
& get\_answer & \textit{Input:} $\rightarrow$ \textit{Output:} result(, evidence) \\
& & Return the detection answer with optional evidence. \\
\bottomrule
\end{tabular}
}
\caption{Instructions of the toolbox in HaluAgent.}
\label{tab: toolkit}
\end{table*}

\subsection{Data Source}
\label{app: data_source}
We introduce the data sources for trajectory generation here, which include five datasets: HaluEval, WebQA, Ape210K, HumanEval, and WordCnt.

\textbullet~\textbf{HaluEval}~\cite{li2023halueval} is a benchmark for evaluating hallucinations in LLMs across various tasks. 
We select 1,000 samples from the QA subset, with 900 samples used for trajectory generation and 100 samples for testing.

\textbullet~\textbf{WebQA}~\cite{li2016dataset} is a QA dataset collected from the Baidu Zhidao platform. 
This dataset contains a large number of questions and corresponding answers. 
We sample 1,000 questions from the training set of WebQA and generate answers for each question using ChatGPT. Then, human annotators evaluate these answers by comparing them to the correct ones to determine if they contain hallucinations. 
We use 900 questions for generating detection trajectories and reserve 100 samples as the testset.

\textbullet~\textbf{Ape210K}~\cite{zhao2020ape210k} is a large-scale math word problem dataset, containing 210,000 Chinese primary school-level math problems. 
Each problem includes a gold answer and the equations needed to derive the answer. Solving Ape210K requires not only natural language understanding but also commonsense knowledge. 
We select 700 data samples from the training set of Ape210K for mathematical calculation. As with the WebQA processing method, we first use ChatGPT to generate answers for each question, and then human annotators label the responses. 
We reserve 100 data samples as the testset, and use the remaining 600 samples for trajectory generation.

\textbullet~\textbf{HumanEval}~\cite{chen2021evaluating} dataset contains 164 programming problems, including function names, comments, specific implementations, and multiple unit tests. It is widely used to test the code generation capabilities of LLMs. We use CodeLlama-34b-Instruct-hf~\cite{roziere2023code} to generate code for the HumanEval dataset, using 100 samples for trajectory generation and the rest for testing.

\textbullet~\textbf{WordCnt} is a newly constructed dataset, representing conditional text generation tasks in scenarios when users interact with LLMs. Specifically, we create WordCnt by prompting GPT-4 to generate a set of text generation instructions with specific length requirements and the corresponding responses for each instruction. WordCnt consists of 200 samples, with 100 samples used for trajectory generation and 100 samples for testing.

\begin{figure*}[tb]
	\centering
	\includegraphics[width=0.90\textwidth]{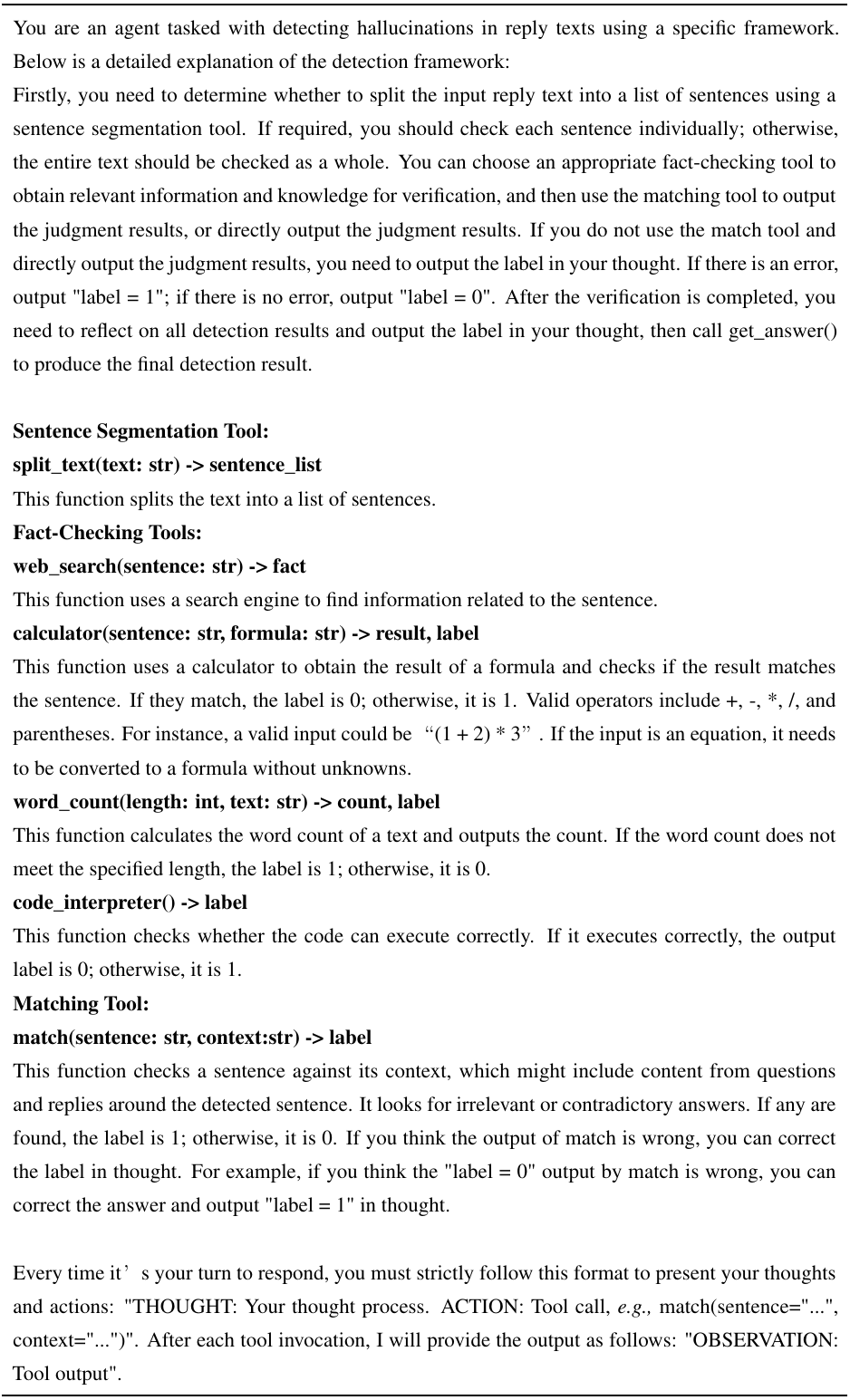}
	\caption{Instructions of HaluAgent framework in English.}
	\label{fig: english_instruction}
	\vspace{-0.2cm}
\end{figure*}

\begin{figure*}[tb]
	\centering
	\includegraphics[width=0.90\textwidth]{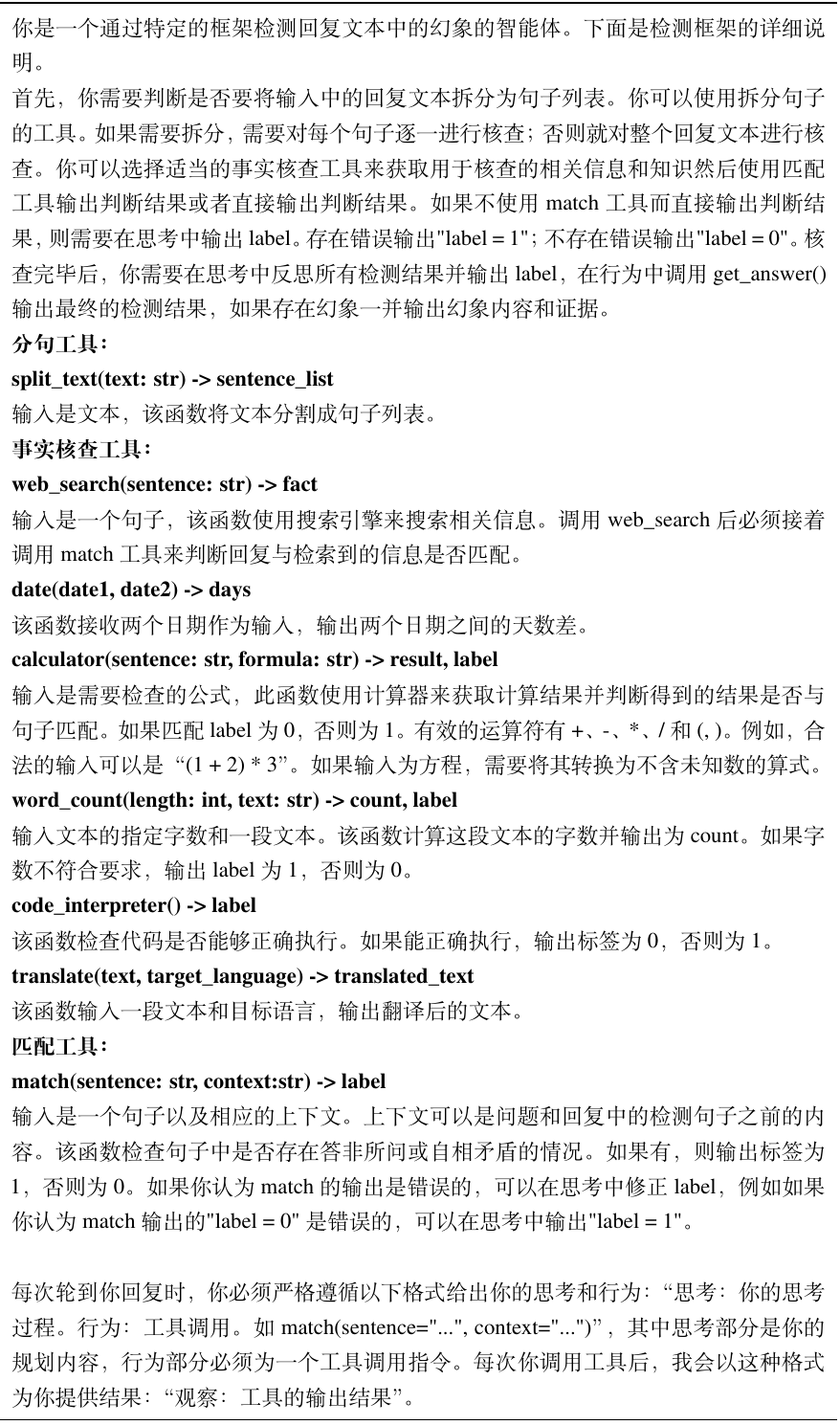}
	\caption{Instructions of HaluAgent framework in Chinese.}
	\label{fig: chinese_instruction}
	\vspace{-0.2cm}
\end{figure*}

\section{Experiment Setup}

\subsection{Out-of-domain Datasets}
\label{app: ood_dataset}
We present the details of the out-of-domain datasets here.

\textbullet~\textbf{HalluQA} is a Chinese Hallucination Question-Answering benchmark covering misleading questions like identity awareness and knowledge-based questions. Each question includes one correct answer and several answers with hallucinations. We conduct experiments on 206 knowledge-related samples from the dataset, randomly selecting one answer from the correct answer and hallucinated answers for evaluation.

\textbullet~\textbf{HaluEval 2.0} is a hallucination evaluation benchmark that contains large-scale questions from five domains: biomedicine, finance, science, education, and open domain. We evenly sample 100 of HaluEval 2.0 for hallucination detection evaluation.

\subsection{Baselines}
\label{app: baselines}
In the experiment section, we compare the hallucination detection performance of GPT-4, Baichuan2-Chat, and HaluAgent. The detailed baseline settings are explained below.

\begin{figure*}[tb]
	\centering
	\includegraphics[width=0.90\textwidth]{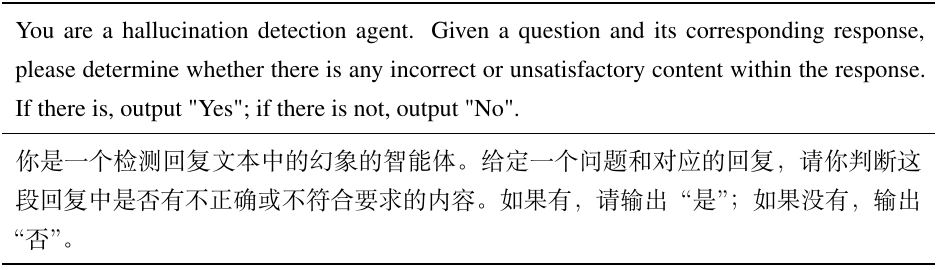}
	\caption{Simple description of the hallucination detection task in English and Chinese.}
	\label{fig: simple_prompt}
	\vspace{-0.2cm}
\end{figure*}

\textbullet~\textbf{GPT-4 prompt} involves providing GPT-4 with a simple description of the hallucination detection task, enabling the model to determine whether there are hallucinations in the text. The model's response is either ``Yes'' or ``No'', indicating the presence or absence of hallucinations. The detailed prompt is shown in Figure~\ref{fig: simple_prompt}.

\textbullet~\textbf{GPT-4 pipeline} guides GPT-4 through the hallucination detection process following the HaluAgent framework, which includes steps such as sentence segmentation and tool invocation. In addition to providing a ``Yes'' or ``No'' answer, it identifies the location of the hallucinations and provides supporting evidence. The instructions are shown in Figure~\ref{fig: english_instruction} and Figure~\ref{fig: chinese_instruction}.

\textbullet~\textbf{Baichuan2-Chat (7B and 13B)} do not have the capability to follow the HaluAgent detection framework. Therefore, we evaluate these models using the same simple hallucination detection prompt as used for GPT-4.

\textbullet~\textbf{HaluAgent (7B and 13B)} are models fine-tuned with trajectory data. We evaluate them using HaluAgent instructions in a zero-shot setting, similar to the evaluation of the GPT-4 pipeline.

\subsection{Implementation Details of Scalability Study}
\label{app: scalability_exp}
\subsubsection{Constructed Dataset}
We construct a new dataset for translation and data calculation tasks via ChatGPT by providing examples. Our goal is to create a dataset specifically for evaluating these new tools, so the questions in the dataset are straightforward.  We present some examples below:
\begin{center}
\textbullet~\textit{Translate the following Spanish into Chinese: ¡Hola! ¿Cómo estás?}

\textbullet~\textit{How many days are there from 2014-02-06 to 2014-05-21?}
\end{center}

\subsubsection{Instructions of New Tools}
\begin{figure*}[tb]
	\centering
	\includegraphics[width=0.90\textwidth]{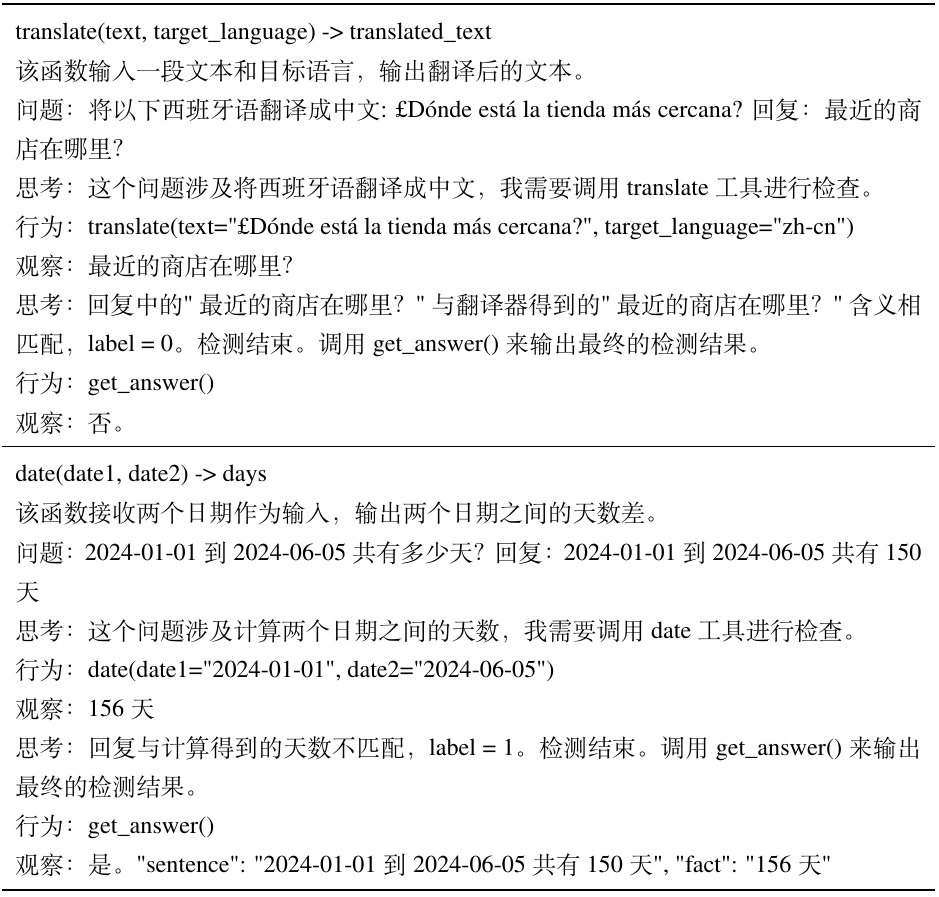}
	\caption{Description and usage example of new tools.}
	\label{fig: new tool prompt}
	\vspace{-0.2cm}
\end{figure*}

To guide HaluAgent in using the new tools, we include descriptions and usage examples of these tools in the instructions. The detailed prompt is shown in Figure~\ref{fig: new tool prompt}.

\end{document}